\documentclass[letterpaper, 10 pt, conference]{utility/ieeeconf} 

\IEEEoverridecommandlockouts  
\usepackage{subfloat}
\usepackage[caption=false,font=small,labelfont=sf,textfont=sf]{subfig}
\usepackage{amssymb}
\usepackage{amsmath}
\usepackage{graphicx} % Required for inserting images
\usepackage{mathtools}
\usepackage{bm}
\usepackage{dsfont}
\usepackage{multirow}
\usepackage{xcolor}
\usepackage{algorithm} 
\usepackage{algorithmic} % or \usepackage{algorithmicx}
\usepackage{url} 
\usepackage[font=footnotesize,labelfont=bf]{caption}
\usepackage{caption}
\usepackage{booktabs}
\usepackage{hyperref} 
\usepackage{cite}  

\title{\LARGE \bf
Diffusion Policies with Value-Conditional Optimization \\ for Offline Reinforcement Learning
}
\author{
 \textcolor{black}{Yunchang~Ma$^{1*}$, Tenglong~Liu$^{1*}$, Yixing~Lan$^{1}$, Xin~Yin$^{1}$, Changxin~Zhang$^{1}$,} \\ \textcolor{black}{Xinglong~Zhang$^{1}$,  Xin~Xu$^{1}$,~\IEEEmembership{Senior Member,~IEEE}}
\thanks{\textcolor{black}{This work was supported by the National Natural Science Foundation of China under Grant 62403483, Grant U24A20279, and Grant U21A20518.
}
}
\thanks{$^{1}\,$Yunchang~Ma, Tenglong~Liu, Yixing~Lan, Xin~Yin, Changxin~Zhang, Xinglong~Zhang and Xin~Xu with College of Intelligence Science and Technology, National University of Defense Technology, Changsha, China. 
(Corresponding author: Yixing~Lan, {\tt\small  lanyixing16@nudt.edu.cn.})}%
\thanks{$^{*}\,$Equal contribution}
}

\begin{document}
\maketitle

%--------------------------------%
\begin{abstract}
    % New Abstract
    In offline reinforcement learning, value overestimation caused by out-of-distribution (OOD) actions significantly limits policy performance. Recently, diffusion models have been leveraged for their strong distribution-matching capabilities, enforcing conservatism through behavior policy constraints. However, existing methods often apply indiscriminate regularization to redundant actions in low-quality datasets, resulting in excessive conservatism and an imbalance between the expressiveness and efficiency of diffusion modeling.
    To address these issues, we propose DIffusion policies with Value-conditional Optimization (DIVO), a novel approach that leverages diffusion models to generate high-quality, broadly covered in-distribution state-action samples while facilitating efficient policy improvement. Specifically, DIVO introduces a binary-weighted mechanism that utilizes the advantage values of actions in the offline dataset to guide diffusion model training. This enables a more precise alignment with the dataset’s distribution while selectively expanding the boundaries of high-advantage actions. During policy improvement, DIVO dynamically filters high-return-potential actions from the diffusion model, effectively guiding the learned policy toward better performance. This approach achieves a critical balance between conservatism and explorability in offline RL.
    We evaluate DIVO on the D4RL benchmark and compare it against state-of-the-art baselines. Empirical results demonstrate that DIVO achieves superior performance, delivering significant improvements in average returns across locomotion tasks and outperforming existing methods in the challenging AntMaze domain, where sparse rewards pose a major difficulty.
    
\end{abstract}
%--------------------------------%

\section{Introduction}
    %----------------------------
    Reinforcement learning (RL) has demonstrated significant success across an extensive array of domains through extensive interactions with environments, including humanoid control~\cite{he2025asap, huang2025learning}, video games~\cite{perolat2022mastering, AlphaStar}, robotics~\cite{kaufmann2023champion, hwangbo2019learning, andrychowicz2020learning, rajeswaran2017learning}, and chess~\cite{silver2016mastering, schrittwieser2020mastering, li2023jiangjun}. However, the traditional trial-and-error approach becomes impractical in a variety of real-world scenarios, such as autonomous driving~\cite{rl4ad}, robot manipulation~\cite{rl4robot}, and healthcare~\cite{rl4healthcare}, where the costs and risks of online data collection are prohibitively high. To address these challenges, offline RL has arisen as a viable option. This approach allows agents to learn from fixed datasets without the need for real-time interaction with the environment~\cite{levine2020offline, BCQ}. Offline RL is particularly advantageous in settings where large, pre-collected datasets are available, offering a practical solution to the limitations of traditional RL methods.
    %----------------------------
    Although offline RL provides a new way to train the learned policy, there is a new challenge in offline RL.
    The main challenge in offline RL is solving the overestimation of the Q-value function caused by out-of-distribution~(OOD) actions. Current methods usually combine the behavior policy generating the fixed dataset to mitigate the OOD problem, which enforces the learned policy to close the behavior policy. 
    
    Policy regularization methods can be broadly classified into two categories: implicit policy regularization and explicit behavior policy regularization.
    Implicit policy regularization methods directly use samples from the fixed dataset to restrict the acquired policy. These approaches typically introduce an additional term that measures the difference between the behavior policy and the trained policy. Common divergence metrics include behavior cloning~\cite{td3bc, td3bc_revisiting}, Fisher divergence~\cite{Fisher-BRC}, Kullback-Leibler divergence~\cite{wu2019behavior, jaques2019way}, and Maximum Mean Discrepancy (MMD)~\cite{BEAR}. While these methods utilize the full fixed dataset to guide policy learning, they may lead to suboptimal solutions, particularly when the dataset quality is poor, as the acquired policy is restricted to closely follow the dataset’s distribution.
    
    In contrast to implicit policy regularization methods, explicit policy regularization techniques model the behavior policy using generative models, such as Variational Autoencoders (VAE)\cite{sohn2015learning} and diffusion models\cite{ho2020denoising}. Several methods utilize VAEs to estimate the behavior policy from offline datasets, including BCQ~\cite{BCQ}, A2PR~\cite{A2PR}, SPOT~\cite{SPOT}, and PLAS~\cite{zhou2021plas}. However, these methods often struggle when the data distribution is complicated, particularly when offline datasets are collected from a variety of policies and exhibit strong multi-modalities~\cite{shafiullah2022behavior}. This challenge highlights the need for more expressive generative models. Recently, the diffusion model’s powerful distribution matching capabilities~\cite{diffusion-ql} have spurred advancements in methods that use diffusion models to represent learned policies~\cite{diffusion-ql, hansen2023idql, sfbc}. Despite these advancements, such approaches remain overly conservative, as they require the learned policy to closely mimic the behavior policy, even when those actions are suboptimal.  Unlike Gaussian methods, diffusion-based approaches require time backpropagation for Q-value loss, introducing instability and computational burden that can harm performance.

    To address the challenges outlined above, we introduce a \textbf{DI}ffusion policies with \textbf{V}alue-conditional \textbf{O}ptimization~(DIVO) method for Offline Reinforcement Learning. Unlike previous approaches that use diffusion models to represent the learned policy, DIVO models the behavior policy using the diffusion model. Additionally, to fully harness the expressive power of the diffusion model and mitigate the impact of low-quality data in the fixed dataset, DIVO integrates the value function with the diffusion model. This integration enhances the model’s ability to capture the distribution of high-quality data in the dataset through a value bootstrapping mechanism, thereby improving the guidance for policy learning and facilitating policy improvement.

    % 与使用diffusion model直接作为学习策略相反，我们将diffusion model作为行为策略，充分发挥其对数据分布的强表达能力，同时通过优势引导机制增强diffusion model对好的数据的建模，从而更好的引导策略提升。

   The paper's contributions can be summarized as follows:
    
    \begin{enumerate}
        \item We propose a diffusion-based behavior policy learning method that selectively expands high-value actions using binary advantage weights while maintaining alignment with the offline dataset. This approach encourages diffusion models to better capture the distribution of high-value regions.
        \item We propose a novel policy optimization method that dynamically filters high-return-potential actions generated by the diffusion model. This enables efficient policy optimization by mitigating the impact of suboptimal data, leading to superior performance across diverse offline RL tasks.
        \item The experiments on the D4RL benchmark demonstrate that DIVO achieves state-of-the-art performance across diverse tasks, significantly improving average returns in locomotion tasks and excelling in challenging sparse-reward scenarios like AntMaze.
        % \item We propose a value-conditional optimization framework grounded in diffusion policies, which establishesdifferentiable policy optimization pathways to achieve synergistic coordination between behavioral conservatism and policy improvement objectives. This framework fundamentally resolves the zero-sum trade-off observed in prior methods that strictly enforce behavioral imitation at the expense of policy performance enhancement.
        % \item We pioneer the integration of the reverse generative process of diffusion models with Q-value function estimation, developing an energy-guided diffusion policy optimization algorithm. This innovation establishes a novel methodological framework for applying generative models in offline reinforcement learning, significantly enhancing the stability and sample efficiency of policy optimization.
    \end{enumerate} % <------------------
\section{Preliminary}

\subsection{Offline RL}

% The environment in RL is typically defined by a Markov Decision Process~(MDP), denoted as $\mathcal M = \{\mathcal S, \mathcal A,  P,  R,\gamma, d\}$. $\mathcal S$ denotes state space, $\mathcal A$ denotes action space, $P(s'|s,a)$ denotes transition probability function, which measures the transition probability from state $s$ to state $s'$ after taking action $a$, $r(s,a)$ denotes the reward function for state-action pair $(s,a)$. $\gamma$ represents the discount. Offline RL aims to find a policy $\pi: \mathcal{S}\rightarrow \Delta(\mathcal{A})$ (or $\pi: \mathcal{S}\rightarrow \mathcal{A}$ if deterministic) from the fixed dataset generated by behavior policies, which maximizes the discounted returns 
In reinforcement learning (RL), the environment is typically modeled as a Markov Decision Process (MDP), denoted as $\mathcal{M} = {\mathcal{S}, \mathcal{A}, P, R, \gamma, d}$. Here, $\mathcal{S}$ represents the state space, $\mathcal{A}$ the action space, and $P(s’|s,a)$ the transition probability function, which defines the probability of transitioning from state $s$ to state $s’$ after taking action $a$. The reward function is denoted by $r(s,a)$, which specifies the reward for a given state-action pair $(s,a)$, while $\gamma$ represents the discount factor. In offline RL, the goal is to identify a policy $\pi: \mathcal{S} \rightarrow \Delta(\mathcal{A})$ (or $\pi: \mathcal{S} \rightarrow \mathcal{A}$ for deterministic policies) from a fixed dataset generated by behavior policies that maximizes the discounted return, expressed as:$J(\pi)=\mathbb{E}_{\pi}\left[\sum_{t=0}^{\infty}\gamma^t r(s_t,a_t)\right]$. Then, the optimal policy $\pi^*$ is then given by: $\pi^*=\arg\max \mathbb{E}_{\pi}\left[\sum_{t=0}^{\infty}\gamma^t r(s_t,a_t)\right]$. For a given initial state $s$, the value function $V(s)$ is defined as: $V(s) = \mathbb{E}_{\pi}\left[\sum_{t=0}^{\infty}\gamma^t r(s_t,a_t)|s_0=s\right]$. Similarly, for an initial state-action pair $(s,a)$, the Q-value function $Q(s,a)$ is given by: $Q(s,a) = \mathbb{E}_{\pi}\left[\sum_{t=0}^{\infty}\gamma^t r(s_t,a_t)|s_0=s,a_0=a\right]$.

Offline RL, unlike traditional RL, does not learn from interactions with the environment. Instead, it aims to learn an optimal policy from a pre-collected dataset $\mathcal{D} = {(s_t, a_t, s_{t+1}, r_t)}$. Offline RL algorithms for continuous control typically follow an actor-critic architecture, which consists of policy evaluation and policy improvement. In the policy evaluation step, a parameterized Q-value function $Q_\phi(s, a)$ is optimized by minimizing the following temporal difference (TD) error:
\begin{equation}\label{eq:td_error}
    \mathcal{L}_{\text {TD}}(\phi) = \mathbb{E}_{\substack{(s,a,s')\in \mathcal{D}, \\ a'\sim \pi_\theta(\cdot|s')}}\left[ r(s,a) + \gamma Q_{\hat{\phi}}(s',a') - Q_{\phi}(s,a) \right],
\end{equation}
where $Q_{\hat{\phi}}(s', a')$ represents the target Q-value function and $\pi_\theta$ is the learned policy.
The value function is used to approximate an expectile based solely on the Q-function, resulting in the following loss function:

\begin{equation}\label{eq:value_update} 
\mathcal{L}_V(\psi) = \mathbb{E}_{(s,a) \sim \mathcal{D}} \left[ (Q_{\phi_i}(s,a) - V_\psi(s))^2 \right], 
\end{equation}
where $V_\psi$ denotes the value function.
\subsection{Diffusion Policy}
Consider a real action data distribution $q(a)$ and a sample $a_0 \sim q(a)$ drawn from it. The forward diffusion process, defined as a Markov chain, progressively adds Gaussian noise to the sample over $K$ steps, following a predefined variance schedule $\beta_k$~\cite{ho2020denoising, song2019generative}, expressed as
\begin{equation}
\begin{aligned}
    q(a_{1:K} | a_0) &= \prod_{k=1}^{K} q(a_k | a_{k-1}), \\
    q(a_k | a_{k-1}) &= \mathcal{N} \left( a_k; \sqrt{1 - \beta_k} a_{k-1}, \beta_k I \right).
\end{aligned}
\end{equation}
Diffusion models learn a conditional distribution $p_\theta(a_{k-1} | a_k)$ and generate new samples by reversing the forward diffusion process described above:
\begin{equation}
\begin{aligned}
    p_\theta(a_{0:K}) &= p(a_K) \prod_{k=1}^{K} p_\theta(a_{k-1} | a_k), \\
p_\theta(a_{k-1} | a_k) &= \mathcal{N} \left( a_{k-1}; \mu_\theta(a_k, k), \Sigma_\theta(a_k, k) \right).
\end{aligned}
\end{equation}
Given $a_0$, the noisy sample $a_k$ can be efficiently obtained using the reparameterization trick:
\begin{equation}\label{eq:re-param}
    a_t = \sqrt{\bar{\alpha}_t} a_0 + \sqrt{1 - \bar{\alpha}_t} \epsilon, \quad \epsilon \sim \mathcal{N}(0, I),
\end{equation}
The practical implementation involves the direct prediction of the Gaussian noise, denoted by $\epsilon$, in Equation~(\ref{eq:re-param}) using a neural network, denoted by $\epsilon_\theta(a_k,k)$, with the objective of minimizing the original evidence lower bound loss,
\begin{equation}
\mathcal{L}(\theta) = \mathbb{E}_{\substack{a_0 \sim p(a_0), \\ k \sim \mathcal{U}, \\ \epsilon \sim \mathcal{N}(0, I)}} \left[ \left\| \epsilon - \epsilon_\theta \left( \sqrt{\bar{\alpha}_t} a_0 + \sqrt{1 - \bar{\alpha}_t} \epsilon, k \right) \right\|^2 \right],
\end{equation}
where the pre-collected offline dataset, denoted as $\mathcal{D}$ and generated under the behavior policy $\pi_b$, is accompanied by $\mathcal{U}$, a uniform distribution defined over the discrete set ${1, \dots, K}$.

\section{Method}

In this section, we introduce DIVO, our method for effective data-driven decision-making. Our approach consists of two components: First, we present the Positive Advantage Diffusion learning~(PAD) method in Section~\ref{sec:pad}, which utilizes an expressive diffusion policy to capture high-value action distributions. Second, we detail the Adaptive Diffusion-based Policy Optimization~(ADPO) method in Section~\ref{sec:adpo}, which extracts the enhanced learned policy from the diffusion behavior while ensuring methodological simplicity.

\subsection{Positive Advantage Diffusion Learning}\label{sec:pad}
% In offline RL, there are many methods combined diffusion models. 
Leveraging the generative capabilities of generative models, numerous approaches incorporating VAE have achieved considerable success in offline RL~\cite{zhou2021plas,BCQ,SPOT}. However, VAE's expressiveness is limited, particularly when dealing with multimodal or heterogeneous fixed datasets~\cite{diffusion-ql}. Recently, several diffusion model methods for offline RL have emerged. These approaches employ diffusion policy as the final policy to train the actor; however, balancing the expressiveness and efficiency of the diffusion model remains challenging~\cite{fang2024diffusion}. To address this issue, we propose the Positive Advantage Diffusion Learning (PAD) method for modeling the behavior policy. We represent the behavior policy $\pi_b$ ia the reverse process of a conditional diffusion model as:
\begin{equation}
    \pi_\omega(a | s) = p_\omega(a_{0:K} | s) = \mathcal{N}(a_K; 0, I) \prod_{k=1}^{K} p_\omega(a_{k-1} | a_k, s)
\end{equation}
where $a_K\sim\mathcal{N}(0,1)$, $\pi_\omega$ is parameterized based on DDPM~\cite{ho2020denoising}. $p_\omega(a_{k-1} | a_k, s)$ can be parameterized as a noise prediction model $\epsilon_\omega$ with the covariance matrix fixed as $\Sigma_{\omega}(a_k, k; s) = \beta_k I$ and mean $\mu_\omega$ constructed as $ \mu_{\theta}(a_k, k; s) = \frac{1}{\sqrt{\alpha_k}} \left( a_k - \beta_k \sqrt{1 - \bar{\alpha}_k} \, \epsilon_{\omega}(a_k, k; s) \right)$. 
The action can be sampled from DDPM based on the following equation:
\begin{equation}
    a_{k-1} = \frac{1}{\sqrt{\alpha_k}} \left( a_k - \frac{\beta_k}{\sqrt{1 - \bar{\alpha}_k}}  \, \epsilon_\omega (a_k, k; s) \right) + \sqrt{\beta_k} \, \epsilon,
\end{equation}
where represents drawing actions from $K$ different Gaussian distributions sequentially, $\epsilon \in \mathcal N(0,1)$, $k$ is the reverse timestep from $\{K,...,1\}$. 

The PAD can be optimized with the following equation:
% \begin{equation}
%     \mathcal{L}(\theta) = \mathbb{E}_{i\in\mathcal{U},\epsilon\in\mathcal{N}(0,I),(s,a)\in\mathcal{D}}[||\epsilon - \epsilon_\theta(\sqrt{\Bar{a}_i}a+\sqrt{1-\Bar{a}_i}\epsilon,s,i)||^2]
% \end{equation}
\begin{align}\label{eq:diffusion_loss}
\mathcal{L}_{\text{PAD}}(\omega) = \mathbb{E}_{\substack{k\in\mathcal{U}, \epsilon\in\mathcal{N}(0,I), \\ (s,a)\in\mathcal{D}}} &\left[ f(s,a)\cdot|| \epsilon - \epsilon_\omega(m_k, k; s) ||^2 \right], \\
f(s,a) = \eta \cdot \mathds{1}(&Q_\phi(s,a)-V_\psi(s)),
\end{align}
where $m_k = \sqrt{\Bar{\alpha}_k}a + \sqrt{1-\Bar{\alpha}_k}\epsilon$, $(s,a)\in\mathcal{D}$ are state-action pairs from the offline dataset, $\eta$ is a hyperparameter that controls the strength of the guidance for diffusion models.
PAD employs a novel weighted behavior cloning (BC) approach based on advantage estimation that sharply distinguishes between beneficial and suboptimal actions. Unlike traditional behavior cloning that indiscriminately imitates all dataset actions, or weighted BC approaches that assign non-zero probability to suboptimal actions through continuous weights, PAD can effectively strengthen the distributions of high-advantage actions from the dataset.

\subsection{Adaptive Diffusion-based Policy Optimization}\label{sec:adpo}

Offline RL confronts the fundamental challenge of balancing two competing objectives: policy improvement and policy constraint. 
To address this dual requirement, offline RL methods typically minimize the following actor loss:
\begin{equation}\label{eq:general_policy_op}
    \mathcal{L}_{\pi}(\theta)=\underbrace{\mathds E_{s\in\mathcal{D},a\sim \pi_\theta}[- \lambda Q_\phi(s,a)]}_{\text{policy improvement}} + \underbrace{ \mathcal{L}_{\text{BC}}(\theta)}_{\text{policy constraint}},
\end{equation}
where $\lambda = \frac{m B}{\sum_{s_i,a_i}Q(s_i,a_i)}$, $B$ represents the batch size.
In Equation~(\ref{eq:general_policy_op}), policy optimization aims to maximize the Q-value function while simultaneously minimizing the behavior cloning loss, thereby ensuring that the learned diffusion policies closely align with the offline dataset. However, unlike the Gaussian case, the diffusion-based objective necessitates backpropagation through time in the Q-value loss due to the recursive nature of numerical solvers. This process is often unstable and computationally expensive, which can hinder performance. Moreover, using the diffusion model as the final policy introduces additional computational overhead, commonly encountered during inference in diffusion models.
\setlength{\textfloatsep}{2mm}  % 控制图表与上下文的间距
\setlength{\intextsep}{2mm}
\begin{algorithm}[ht]
    \caption{Diffusion Policies with Value-Conditional Optimization for Offline Reinforcement Learning~(DIVO)}
    \label{pseudo_code}
    \begin{algorithmic}
    \STATE {\bfseries Input:} {$\alpha$: hyper-parameters, $\mathcal{D}$: Replay buffer,   $N$: batch size,  $\tau$: target network update rate .}

    \STATE The Q-value network and the target Q-value network are initialized with parameters  $\phi_1, \phi_2$, policy network is initialized with $\theta$ and value function network is initialized with $\psi$, target Q and target policy network are initialized with $\phi_1^{'} \leftarrow{\phi_1}, \phi_2^{'}\leftarrow{\phi_2}, \theta' \leftarrow{\theta}$, Diffusion networks are initialized with $\epsilon_{\omega}$.

        \FOR{$t=1$ {\bfseries to} $T_1$}
        \STATE  Select a subset of transitions $(s,a,r,s')\sim \mathcal{D}$\\
        \textbf{Positive Advantage Diffusion Learning:}\\
        \quad Optimize by reducing the Equation~(\ref{eq:diffusion_loss})\\
        \textbf{Q-function and value-function update:}\\
        \quad Optimize Q-value through minimizing Equation ~(\ref{eq:td_error})\\
        \quad Optimize Value function through minimizing Equation~(\ref{eq:value_update})\\
        \textbf{Adaptive Diffusion-based Policy Optimization:} \\
        \quad Optimize policy network through minimizing Equation~(\ref{eq:policy_opti})\\
        \textbf{Update Target Networks: }\\
        \qquad $\phi^{'}_i \leftarrow{\tau\phi + (1 - \tau)\phi^{'}_i},i=1,2$ \\
        \qquad $\theta^{'} \leftarrow{\tau\theta + (1 - \tau)\theta^{'}}$
        \ENDFOR
    \end{algorithmic}
\end{algorithm}

To address these challenges, we propose the Adaptive Diffusion-based Policy Optimization (ADPO) method. ADPO leverages high-quality actions from the diffusion model to guide policy learning, fully utilizing the expressive properties of the diffusion model. The final target action $\pi_\text{target}$ is selected based on the advantage of the action sampled from the behavioral diffusion policy, as defined by the following equation:
\begin{equation}\label{final_a}
    \begin{aligned}
                \pi_{\text{target}} &=\left\{
                    \begin{array}{ll}
                      \pi_\omega(s), \quad\quad A(s,\pi_\omega(s)) \geq 0,\\
                      \pi_{\phi}(s),\quad\quad A(s,\pi_\omega(s))< 0,\\
                    \end{array}
                  \right.\\
    \end{aligned}
    \end{equation}
where $\pi_\omega(s)$ represents the action from the behavioral diffusion policy based on the state $s$. It means choosing the action with a high advantage from the diffusion model as the target action, which can be used to constrain the learned policy to achieve policy improvement.

Within ADPO, the diffusion model serves as the behavior policy rather than the learned policy during the policy optimization stage. The final policy optimization objective combines policy improvement and policy regularization:

\begin{equation}\label{eq:policy_opti}
    \mathcal{L}_{\pi}(\theta) = \mathbb E_{s\in\mathcal{D}}\left[-\alpha  Q_\phi(s,\pi_\theta(s)) + \beta ||\pi_\theta(s) - \pi_{\text{target}}(s)||^2\right]
\end{equation}

% \begin{equation}
% \mathcal{L}{\pi}(\theta) = \mathbb{E}{s\in\mathcal{D}}\left[-\alpha Q_\phi(s,\pi_\theta(s)) + \beta |\pi_\theta(s) - \pi_{\text{target}}(s)|^2\right]
% \end{equation}
% Where:
% \begin{equation}
% \pi_{\text{target}}(s) =
% \begin{cases}
% \pi_\omega(s) & \text{if}\ A(s,\tilde{a}) \geq 0 \
% \pi_\phi(s) & \text{if}\ A(s,\tilde{a}) < 0
% \end{cases}
% \end{equation}
% ADPO adaptively selects actions with high value from the diffusion model to constrain the learned policy, which maximizes the Q-value function while ensuring it closely follows the behavior policy. Meanwhile, this helps avoid the instability and high computational cost associated with backpropagating through time in the Q-value loss within the diffusion model. 
where $\beta$ is a hyperparameter that balances policy improvement and policy regularization. DIVO can adaptively select high-advantage actions from the diffusion model to constrain the learned policy, thereby maximizing the Q-value function while ensuring close alignment with the behavior policy. This approach simultaneously avoids the instability and computational overhead associated with backpropagating through time in the Q-value loss within the diffusion model. Meahwhile, we provide the pseudo-code for DIVO, as presented in Table~\ref{pseudo_code}. Our algorithm is built upon the TD3+BC framework~\cite{td3bc}.

% Unlike existing approaches that model behavioral policies, our DIVO framework introduces a fundamentally different perspective. Traditional methods such as behavior cloning (BC) directly imitate the dataset distribution without considering the quality of actions, while weighted BC approaches attempt to address this limitation by incorporating exp(Q-V) as weights in the loss function. However, these approaches still operate within the confines of direct imitation learning.

\begin{figure*}[htbp]
	\centering

	%\vspace{-0.3cm}	

	\subfloat[HalfCheetah]{
		\includegraphics[width=0.45\columnwidth]{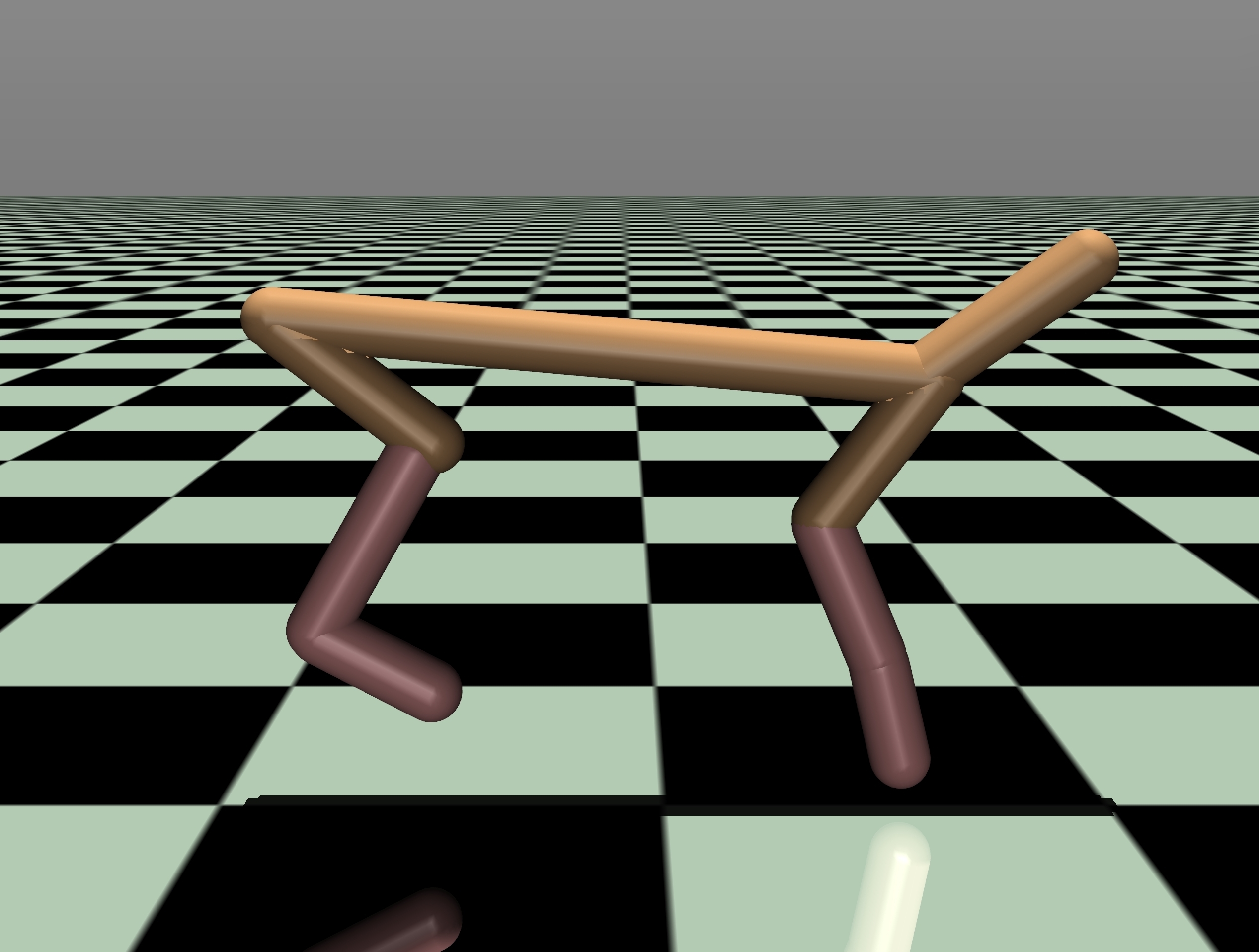}
		\label{fig:halfcheetah_scene}
	}
    \hspace{-6pt}
	\subfloat[Hopper]{
		\includegraphics[width=0.45\columnwidth]{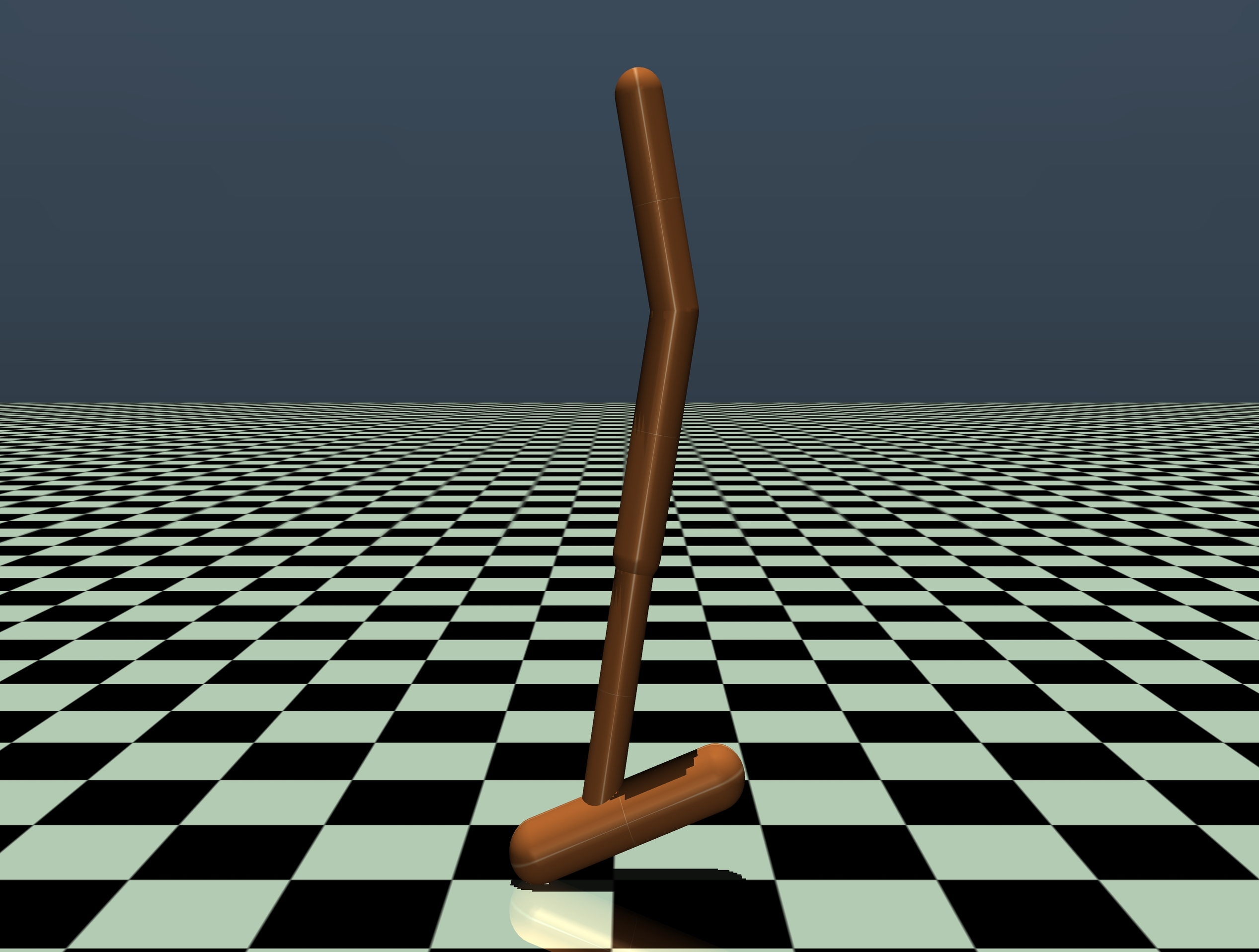}
		\label{fig:hopper_scene}
	}
	%	\hspace{0.1cm}
	\subfloat[Walker2d]{
		\includegraphics[width=0.473\columnwidth]{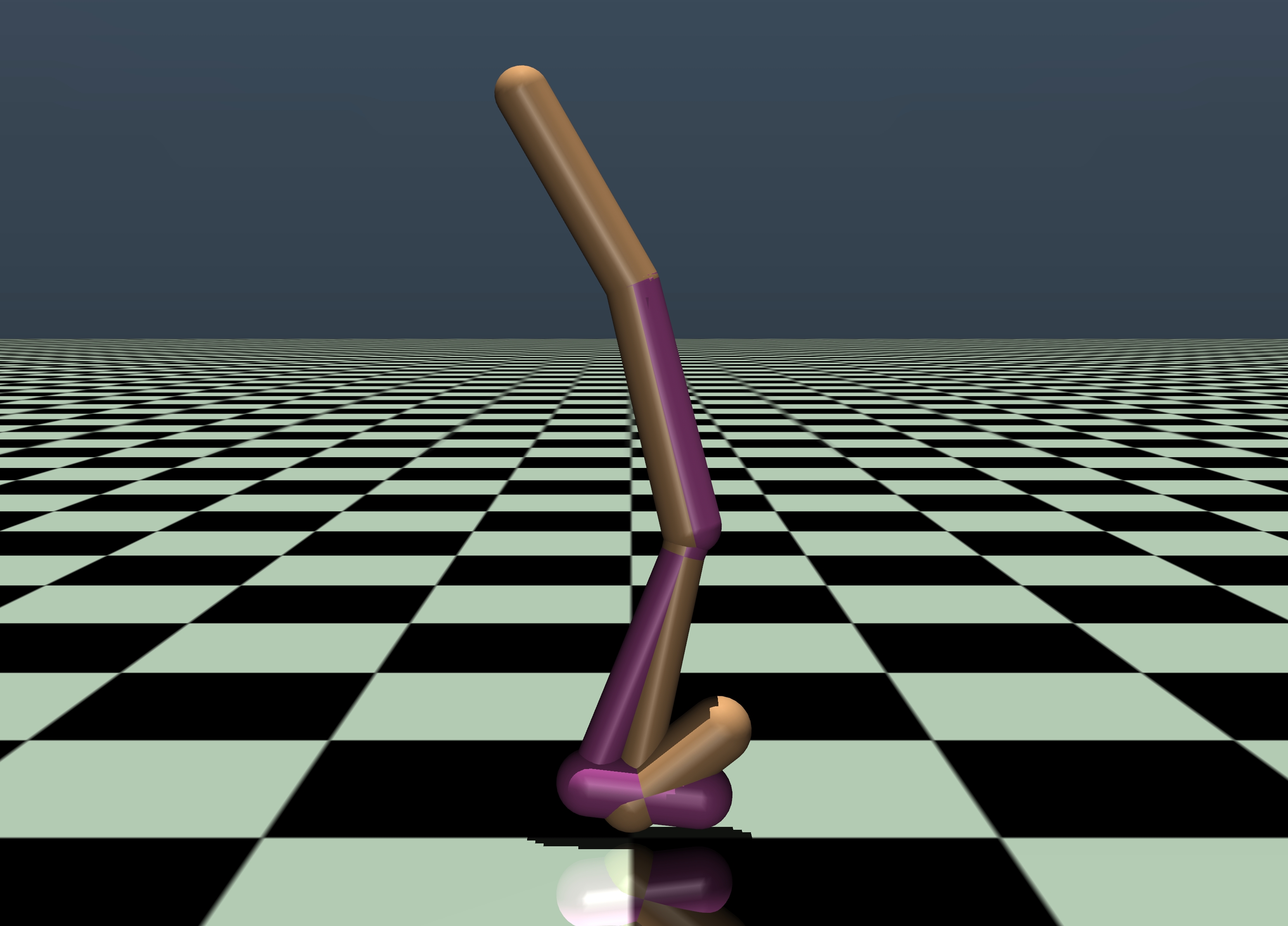}
		\label{fig:walker2d_scene}
	}
        \subfloat[Ant]{
        \includegraphics[width=0.452\columnwidth]{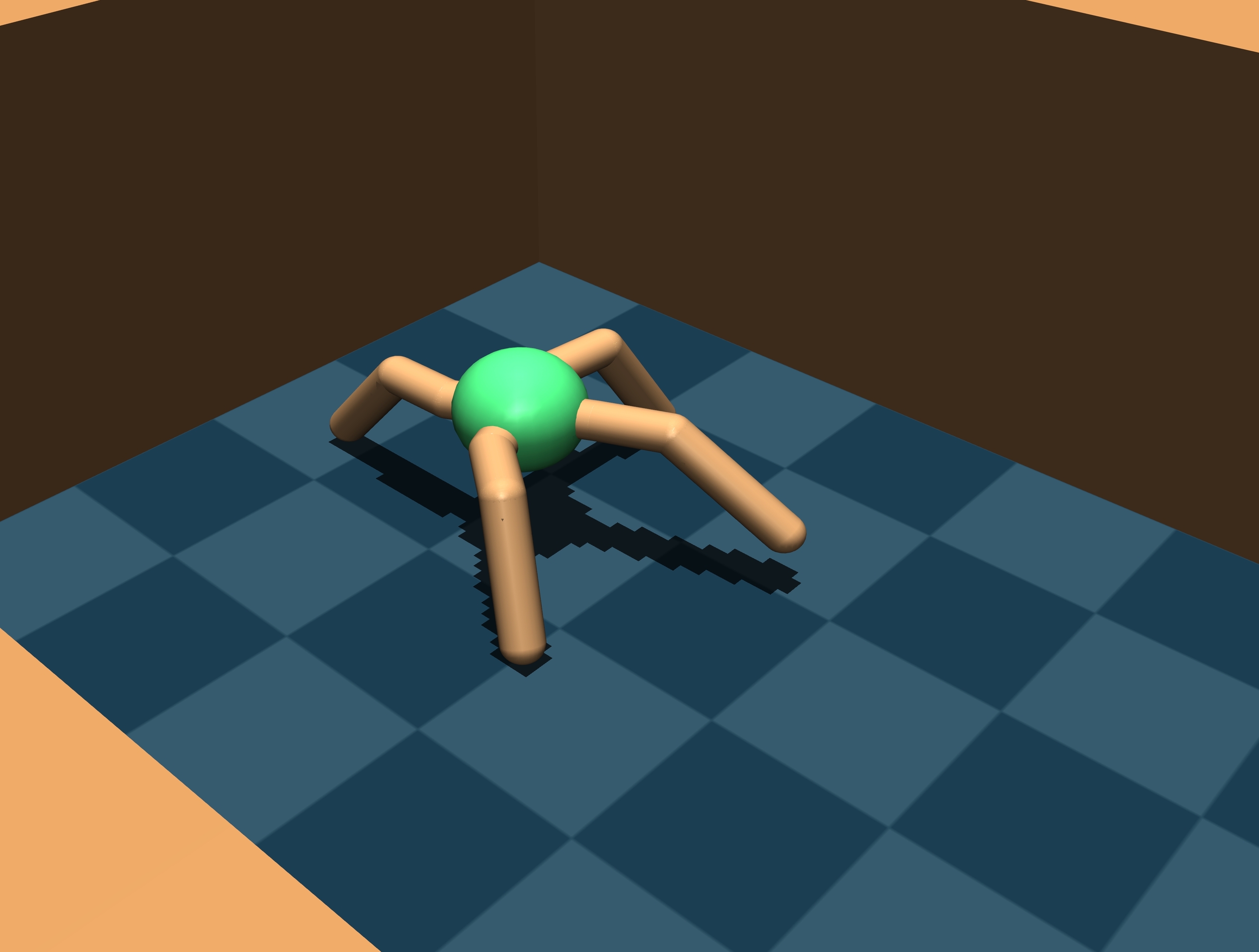}
        \label{fig:ant_scene}
    }	
        
	\caption{
		The robot continuous control offline RL benchmarks including ant, halfcheetah, hopper and walker2d.
	}	
	\label{fig:mujoco}
	%\vspace{-0.5cm}
\end{figure*}

\begin{table*}[t]
    % \vspace{-10pt}
    \centering
    \small
    \setlength{\tabcolsep}{4pt}
    \renewcommand{\arraystretch}{1.3}  
    \begin{tabular}{lcccccccc||c}
    \hline
    \toprule
    \multicolumn{1}{c}{\bf Task Name}  & \multicolumn{1}{c}{\bf TD3+BC} & \multicolumn{1}{c}{\bf BCQ} & \multicolumn{1}{c}{\bf BEAR} & \multicolumn{1}{c}{\bf CQL} & \multicolumn{1}{c}{\bf EDP} & \multicolumn{1}{c}{\bf IDQL} & \multicolumn{1}{c}{\bf Diffusion-QL} &\multicolumn{1}{c}{\bf DTQL}& 
        \multicolumn{1}{c}{\bf DIVO~(Ours)} \\ 
    \midrule
       
        halfcheetah-medium & $48.3$ & $47.0$ & $41.0$  & $44.0$ & $52.1$  &$51.0$& $51.1 \pm 0.5$ &$57.9 \pm 0.13$ & $\bm{67.71 \pm 0.64 }$\\ 
        hopper-medium & $59.3$ & $56.7$ & $51.9$  & $58.5$ & $81.9$  &$65.4$& $90.5 \pm 4.6$ &$ 99.6 \pm 0.87$ & $\bm{100.34 \pm 0.44}$  \\
        walker2d-medium & $83.7$ & $72.6$ & $80.9$  & $72.5$ & $86.9$  &$82.5$& $87.0 \pm 0.9$ & $89.4 \pm 0.13$ & $\bm{90.94\pm 1.63}$ \\ 
        \midrule
        halfcheetah-medium-replay & $44.6$ & $40.4$ & $29.7$ & $45.5$ & $49.4$ &$45.9$ & ${47.8 \pm 0.33}$ & $50.9 \pm 0.11$ & $ \bm{55.53\pm 0.77}$\\ 
        hopper-medium-replay & $60.9$ & $53.3$ & $37.3$ & $95.0$ & $101.1$  &$92.1$& $\bm{101.3 \pm 0.6}$ & $100.0 \pm 0.13$ & $\bm{101.28 \pm 0.40}$\\ 
        walker2d-medium-replay & $81.8$ & $52.1$ & $18.5$  & $77.2$ & $94.9$ &$85.1$ & $\bm{95.5 \pm 1.5}$ &$88.5\pm2.16$ & $94.85 \pm0.57 $\\ 
        \midrule
        halfcheetah-medium-expert & $90.7$ & $89.1$ & $38.9$  & ${91.6}$ & $95.5$  &$95.9$& $96.8 \pm 0.33$ &$92.7\pm0.2$ & $\bm{97.88 \pm0.74}$\\ 
        hopper-medium-expert & $98.0$ & $81.8$ & $17.7$   & ${105.4}$ & $97.4$ &$108.6$ & $\bm{111.1 \pm 1.3}$ &$109.3\pm1.49$ & $\bm{111.02  \pm 1.64}$ \\ 
        walker2d-medium-expert & $110.1$ & $109.5$ & $95.4$  & $108.8$ & $110.2$ &$\bm{112.7}$ & $110.1 \pm 0.33$ &$110.0\pm0.07$ & $\bm{112.03\pm 0.62}$\\ 
            \midrule
            {\bf{Gym Average}} & $677.4$ & $602.5$ & $411.3$ & $698.5$ & $769.5$ & $739.2$& $792.0$ &$798.3$ &$\bm{831.6}$ \\ 
        \midrule
        antmaze-umaze & $91.3$ & $0.0$ & $73.0$  & $84.8$ & $96.6$ &$94.0$& $93.4\pm3.4$ & $94.8\pm1.00$ & $\bm{100.00\pm0.00}$ \\ 
        antmaze-umaze-diverse & $54.6$ & $61.0$ & $61.0$ & $43.3$ & $69.5$ &$80.2$& $66.2\pm8.6$ & $78.8\pm1.83$ & $\bm{92.00\pm4.47}$\\ 
        antmaze-medium-play & $0.0$ & $0.0$ & $0.0$  & $65.2$ & $0.0$ &$84.5$& $76.6\pm10.8$ & $79.6\pm1.8$ & $\bm{86.70\pm4.31}$ \\ 
        antmaze-medium-diverse & $0.0$ & $0.0$ & $8.0$ & $54.0$ & $6.4$ &$84.8$& $78.6\pm10.3$ & $82.8\pm1.71$ & $\bm{86.70\pm12.47}$\\ 
        antmaze-large-play & $0.0$ & $6.7$ & $0.0$ & $18.8$ & $1.6$ &$63.5$& $46.6\pm8.3$ & $52.0\pm2.23$ & $\bm{58.00\pm8.37}$\\ 
        antmaze-large-diverse & $0.0$ & $2.2$ & $0.0$ & $31.6$ & $4.4$ &$\bm{67.9}$& $56.6\pm7.6$ & $54.0\pm2.23$ & $63.33\pm4.71$ \\ 
		\midrule
		{\bf{Antmaze Average}} & $145.9$ & $69.9$ & $142.0$ & $297.7$ & $178.8$ & $474.6$ &$417.6$ &$441.6$ &$\bm{486.7}$ \\ \midrule
		{\bf{Total Average}} & $823.3$ & $672.4$ & $553.3$ & $996.2$ & $948.3$ & $1213.8$ &$1209.6$ &$1239.9$ &$\bm{1318.3}$ \\ 
		\bottomrule
      \hline
    % \vspace{-.3in}
	\end{tabular}
	% \label{table:d4rl}
     % \vskip -0.1in
         \caption{The effectiveness of DIVO and rival baseline methods on D4RL datasets (Gym, AntMaze). The outcomes for DIVO represent the average and standard errors of normalized D4RL scores across the last 10 evaluations and 5 different random seeds. 
    }
    \label{experi:d4rl-result}
    \vspace{-10pt}
\end{table*}

\begin{figure*}[ht]
% \vskip -0.1in
% \centering
\subfloat{
		\begin{minipage}[t]{0.33\linewidth}
			\centering
			\includegraphics[width=2.2in]{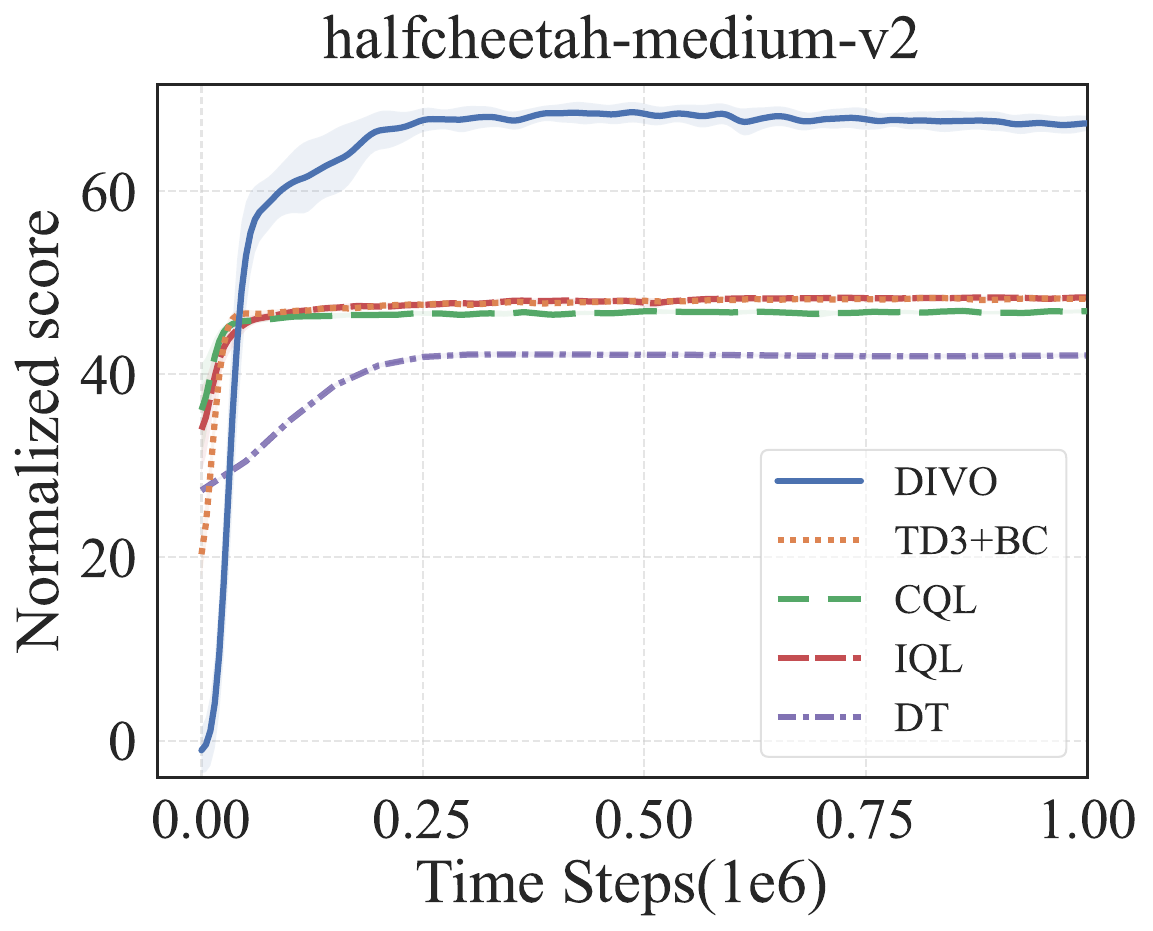}
		\end{minipage}%
	}%
\hspace{-10pt}
\subfloat{
		\begin{minipage}[t]{0.33\linewidth}
			\centering
			\includegraphics[width=2.2in]{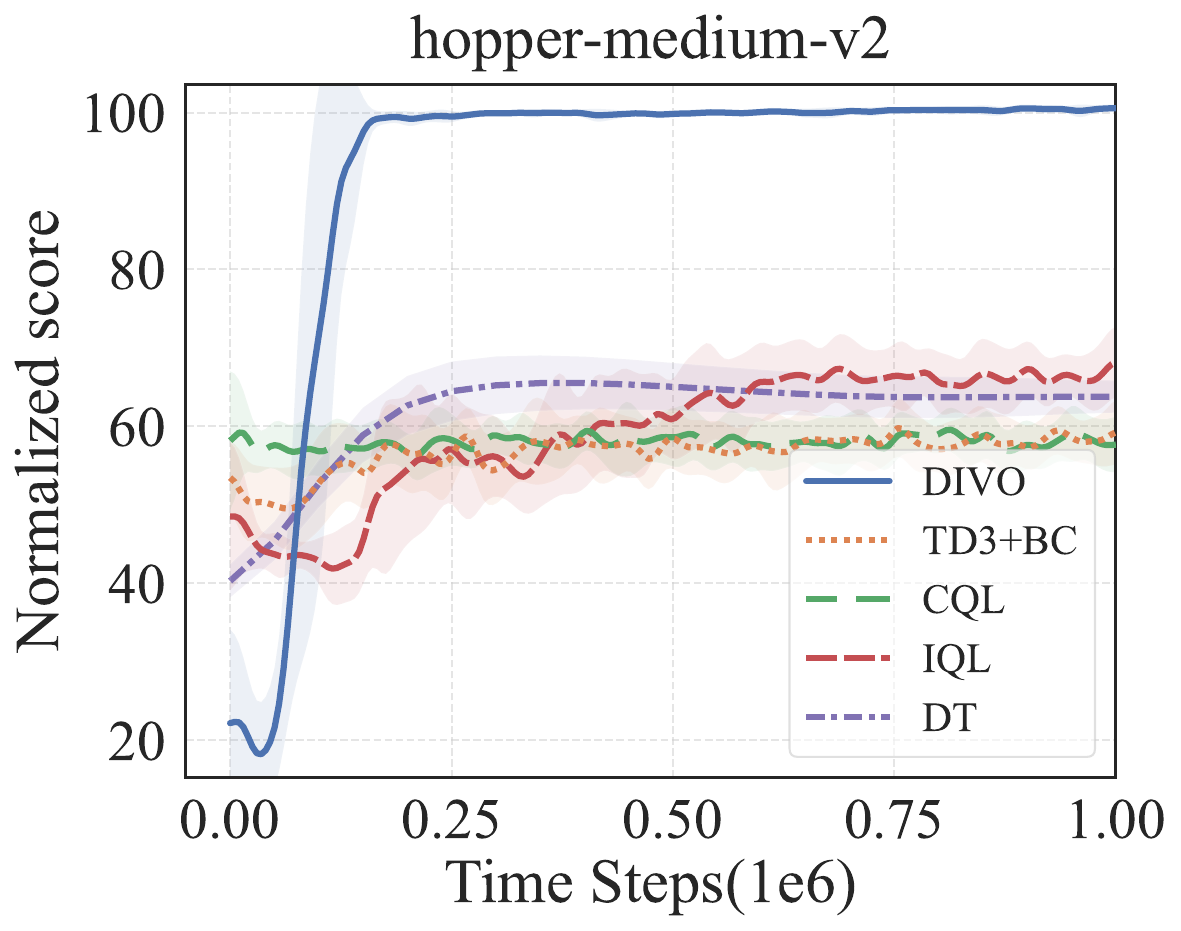}
		\end{minipage}%
	}%
\hspace{-10pt}
\subfloat{
		\begin{minipage}[t]{0.33\linewidth}
			\centering
			\includegraphics[width=2.2in]{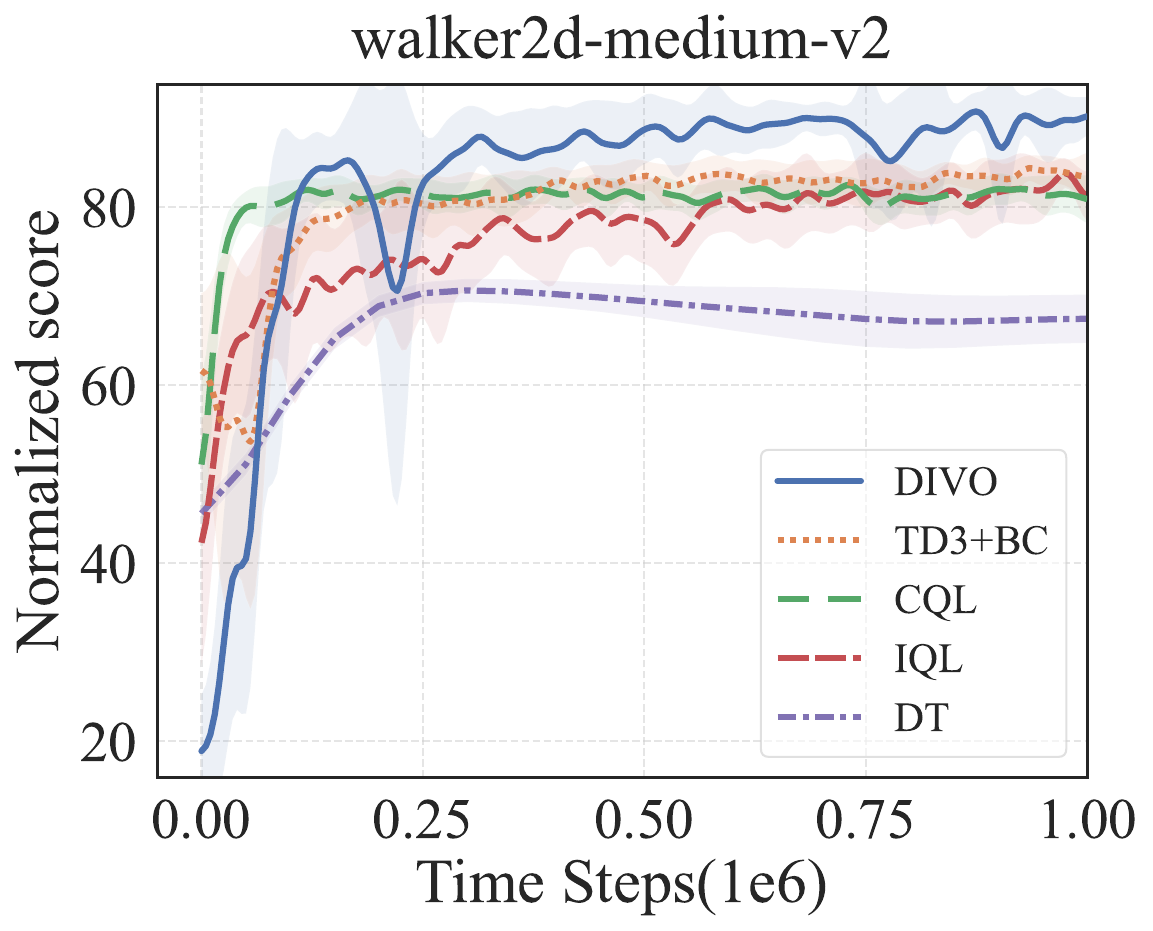}
		\end{minipage}%
}%
\vspace{-.1in}

\subfloat{
		\begin{minipage}[t]{0.33\linewidth}
			\centering
			\includegraphics[width=2.2in]{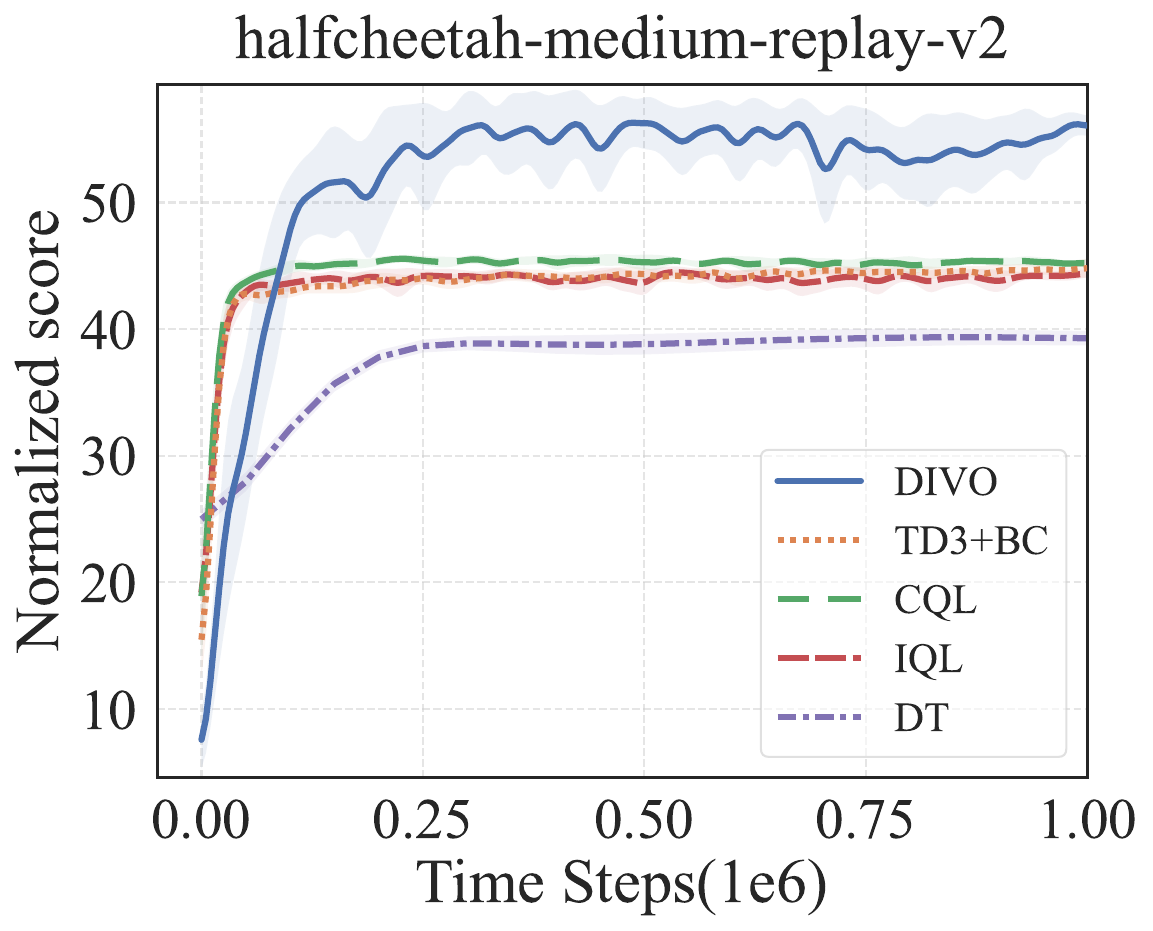}
		\end{minipage}%
	}%
\hspace{-10pt}
\subfloat{
		\begin{minipage}[t]{0.33\linewidth}
			\centering
			\includegraphics[width=2.2in]{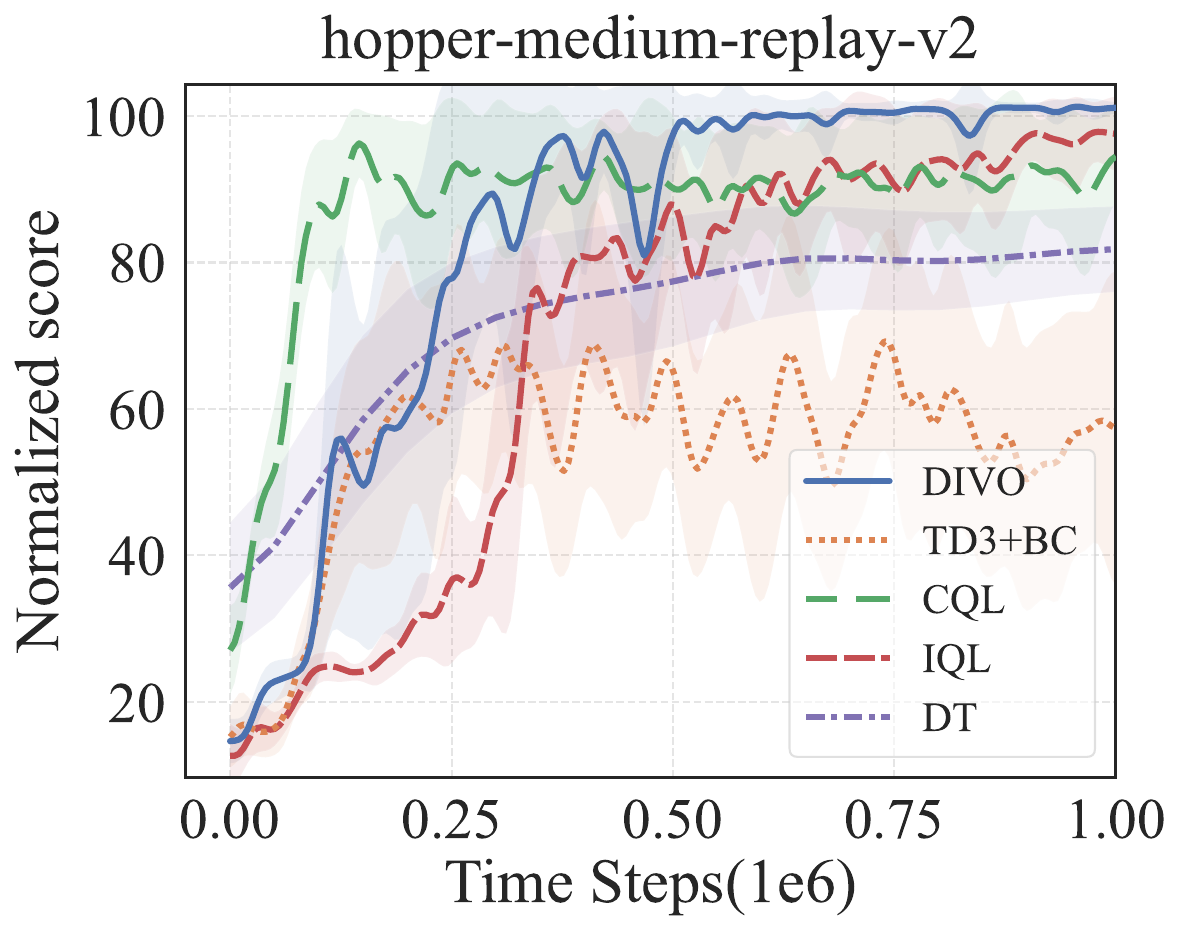}
		\end{minipage}%
	}%
\hspace{-10pt}
\subfloat{
		\begin{minipage}[t]{0.33\linewidth}
			\centering
			\includegraphics[width=2.2in]{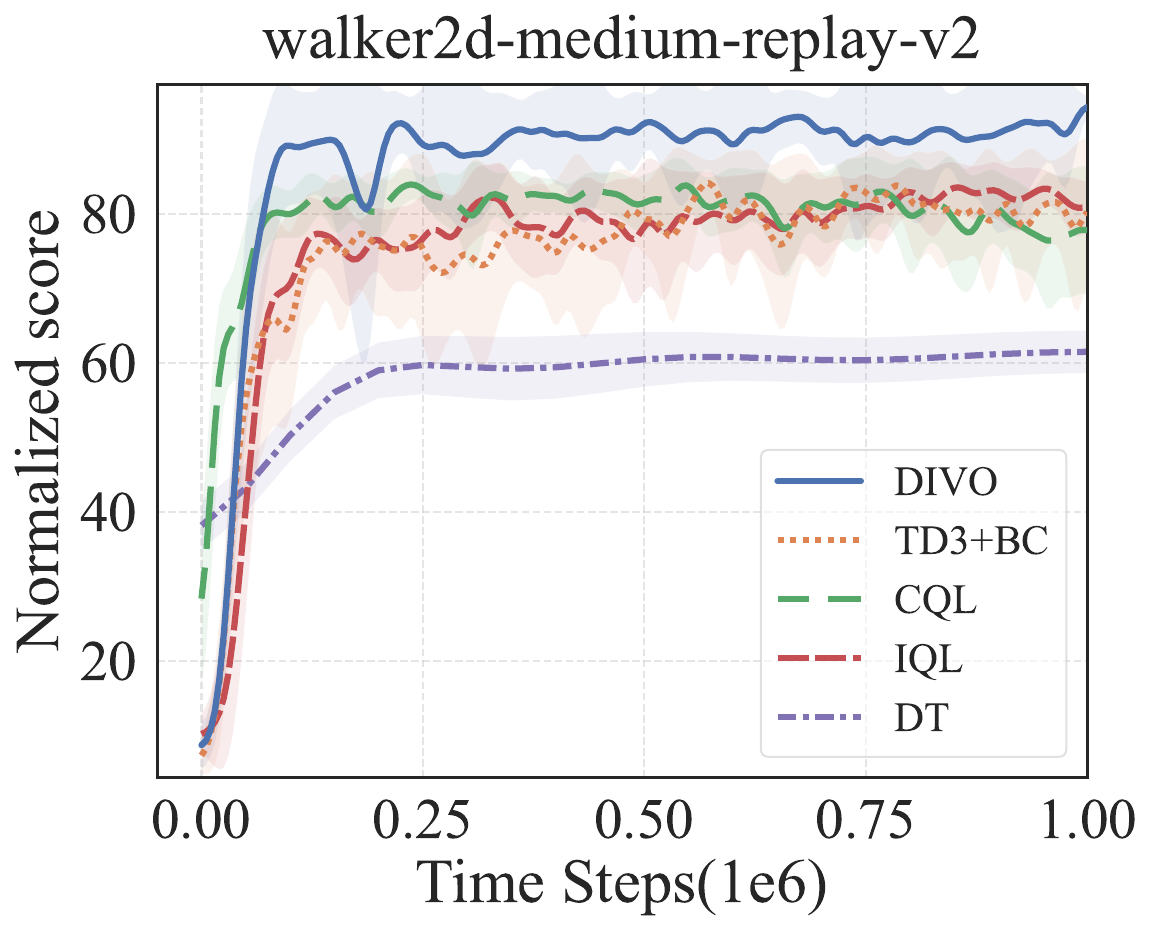}
		\end{minipage}%
	}%
\vspace{-.1in}

\hspace{-5pt}
\subfloat{
		\begin{minipage}[t]{0.33\linewidth}
			\centering
			\includegraphics[width=2.2in]{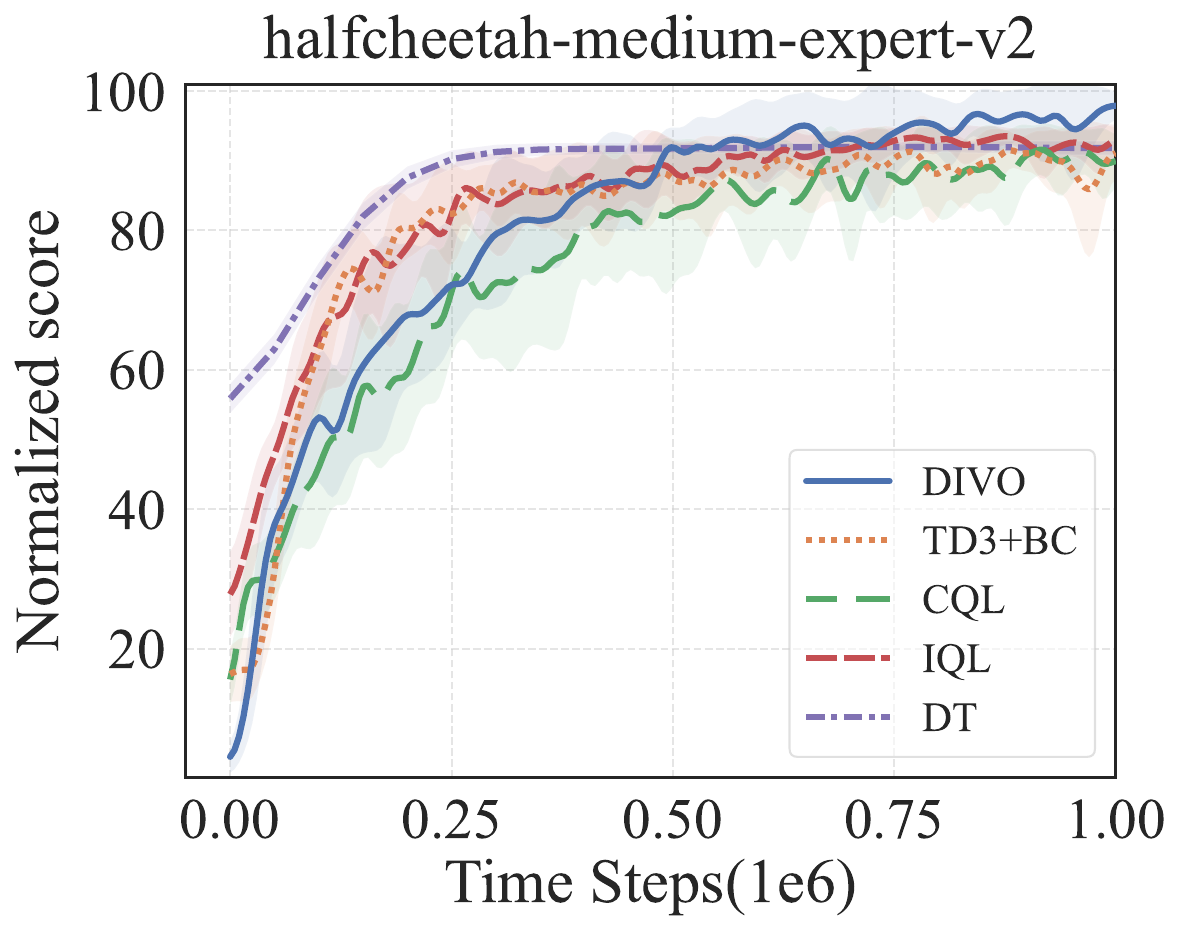}
		\end{minipage}%
	}%
\hspace{-10pt}
\subfloat{
		\begin{minipage}[t]{0.33\linewidth}
			\centering
			\includegraphics[width=2.2in]{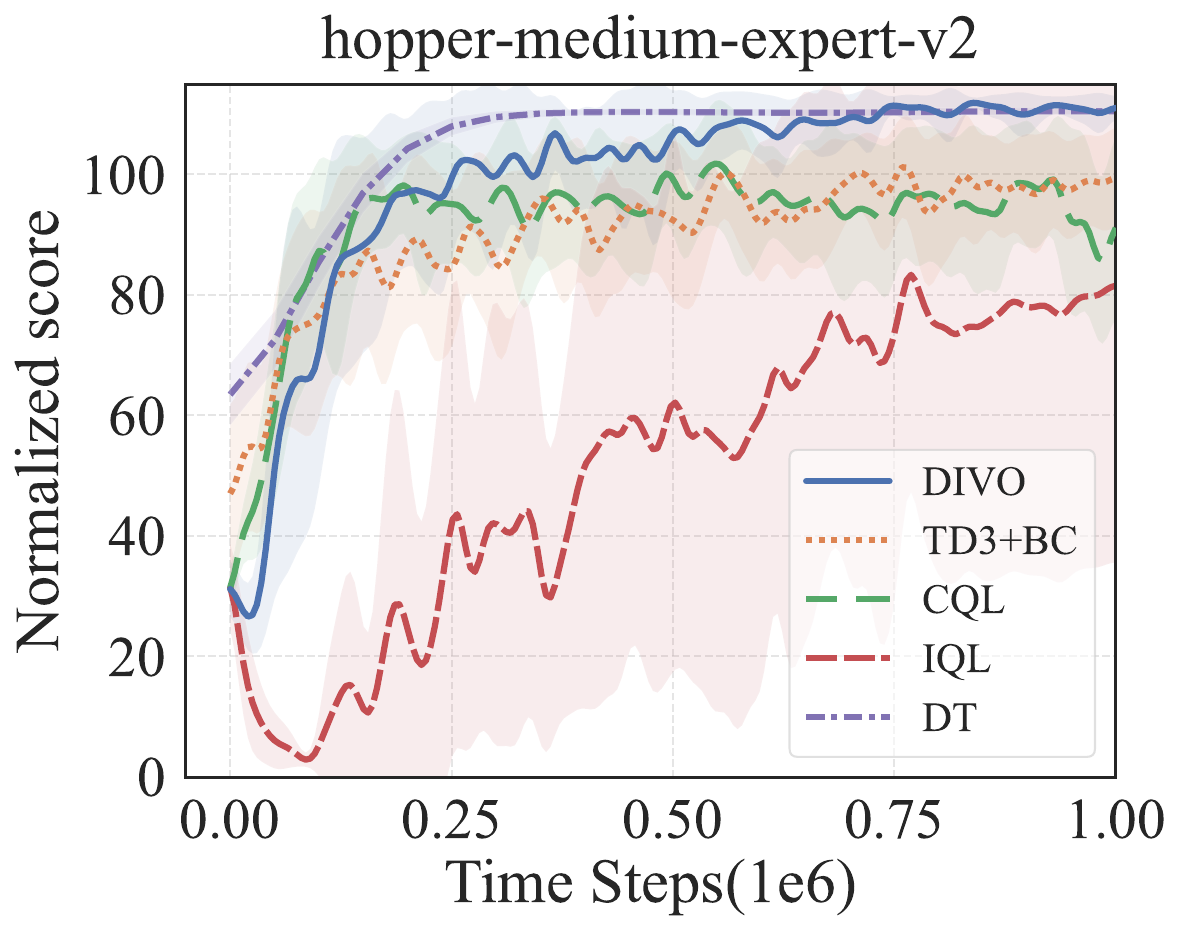}
		\end{minipage}%
	}%
\hspace{-10pt}
\subfloat{
		\begin{minipage}[t]{0.33\linewidth}
			\centering
			\includegraphics[width=2.2in]{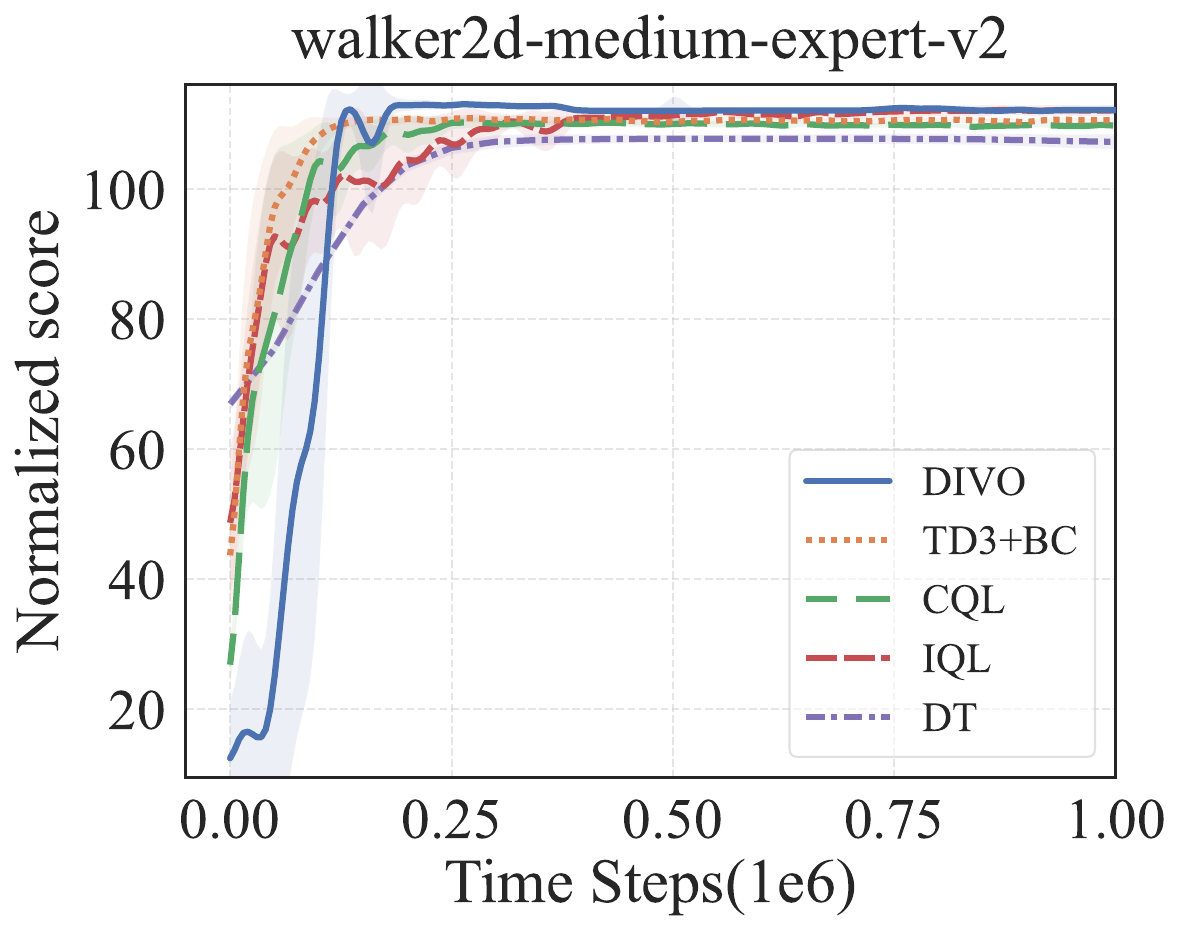}
		\end{minipage}%
	}%
\vspace{.1in}

\centering
% \vspace{-.1in}

\caption{Performance comparison outcomes for nine original tasks within the D4RL dataset. The lines and shaded regions represent the mean values and standard deviations, computed across 5 different random seeds respectively.}
% \vskip -0.1in
% 	\vspace{-0.5cm} 
\label{fig:learning_curve}
\vspace{-10pt}
\end{figure*}

\section{Experiments}
In this section, we start by outlining the experimental setup in Section~\ref{experi: setup}, which includes the description of the offline datasets and the baseline method for our experiments. Subsequently, we introduce the main findings on the D4RL benchmark dataset in Section~\ref{experi:main_res}, which shows that DIVO can achieve the state-of-the-art~(SOTA) performance compared to the strong baseline methods. Finally, in Section~\ref{sec:hyperpara}, we perform a hyperparameter sensitivity analysis to assess the impact of different hyperparameter values on the performance of the DIVO algorithm across the D4RL benchmark.
\begin{figure*}[ht]
	% \centering
	%\vspace{-0.3cm}	
	\subfloat{
		\includegraphics[width=\textwidth]{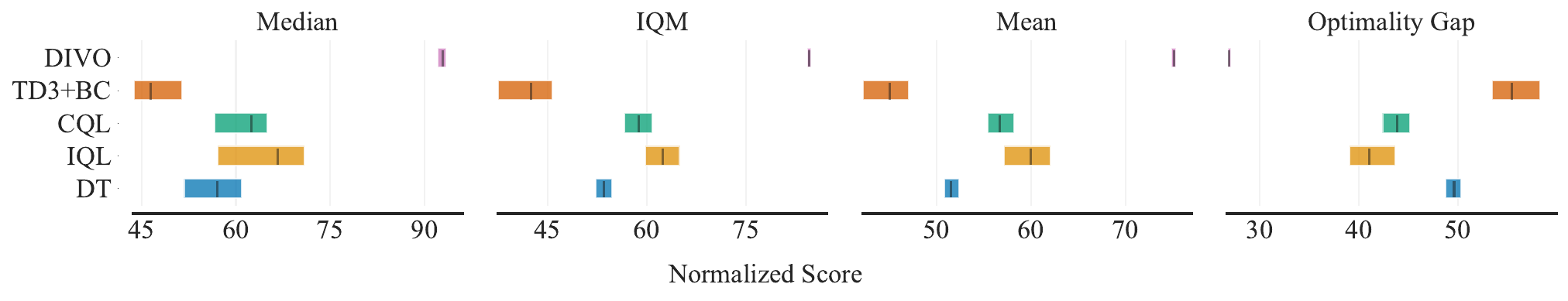}
		\label{fig:halfcheetah_scene}
	}
	\caption{
		Robust assessment of statistical uncertainty on D4RL using 95\% confidence intervals derived from 18 tasks, with 5 random seeds per task.
	}	
	\label{fig:score}
	\vspace{-10pt}
\end{figure*}

% \begin{figure}[ht]
%     % \centering
%         \includegraphics[width=\linewidth]{ieeeconf_IROS/figure/ablation/hopper-medium-v2-eta1.pdf}
%         \caption{ The hyperparameter sensitivity analysis examines the effect of the weight $\eta$ in the optimization objective of the diffusion model.}
%         \label{fig:eta}
%     \end{figure}

% \begin{figure}[ht]
%     % \centering
%         \includegraphics[width=\linewidth]{ieeeconf_IROS/figure/ablation/hopper-medium-v2-beta1.pdf}
%         \caption{ The hyperparameter sensitivity analysis examines the effect of the weight $\beta$ in the optimization objective of policy improvement.}
%         \label{fig:beta}
%         \vspace{-10pt}
%     \end{figure}

\begin{figure}[htbp]
	\centering

	%\vspace{-0.3cm}	

	\subfloat[HalfCheetah]{
		\includegraphics[width=0.47\columnwidth]{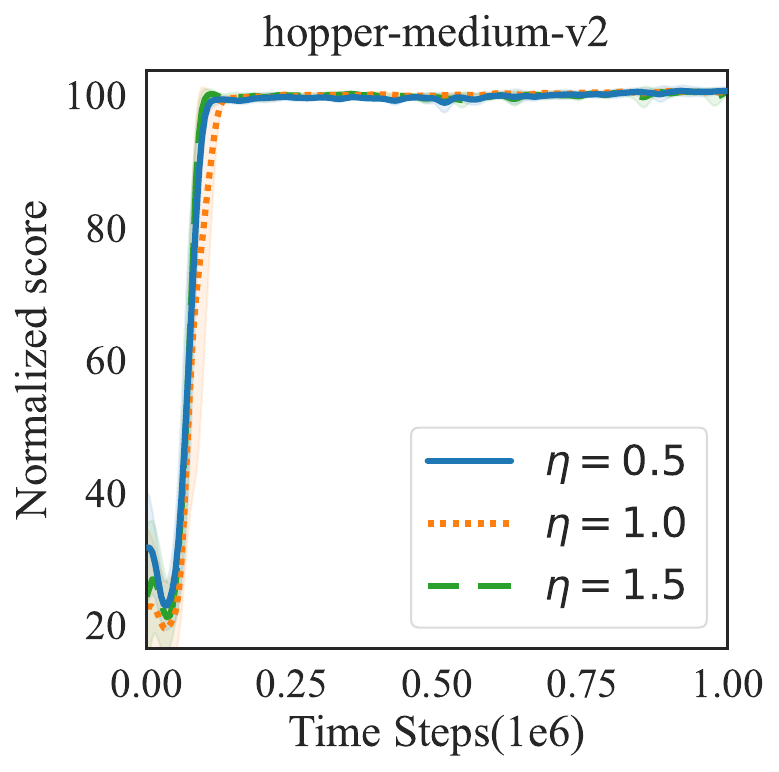}
		\label{fig:halfcheetah_scene}
	}
    \hspace{-6pt}
	\subfloat[Hopper]{
		\includegraphics[width=0.47\columnwidth]{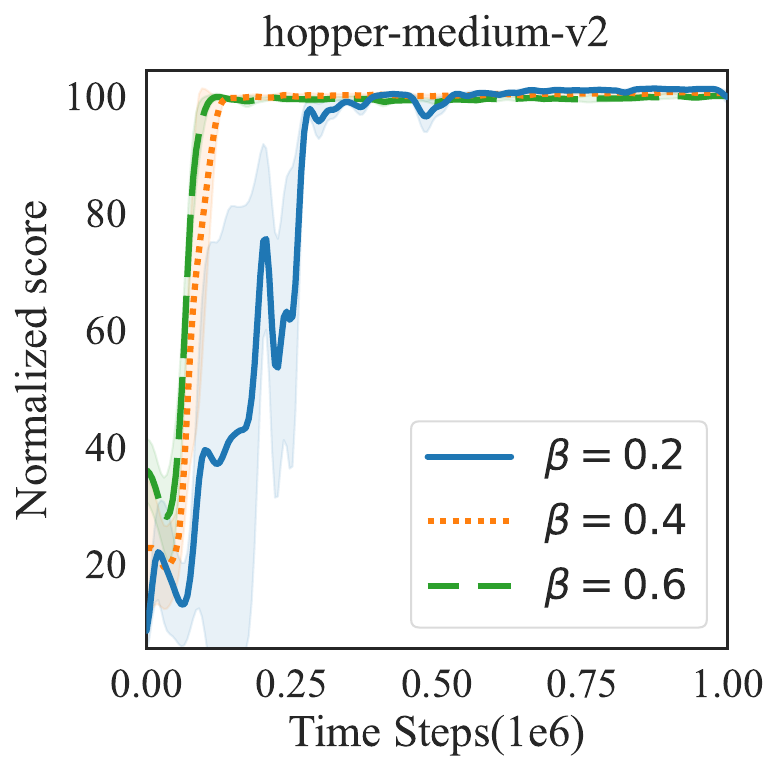}
		\label{fig:hopper_scene}
	}
        
	\caption{
		The hyperparameter sensitivity analysis examines the effect of the weight $\eta$ in the optimization objective of the diffusion model and the weight $\beta$ in the optimization objective of policy improvement.
	}	
	\label{fig:hyperany}
	%\vspace{-0.5cm}
\end{figure}

\subsection{Setup}\label{experi: setup}
\paragraph{Datasets}
 Our experimental evaluations are conducted on two distinct task domains from the D4RL benchmark~\cite{d4rl} using v2 standardized datasets: the Gym locomotion suite and the AntMaze navigation environment. The Gym-MuJoCo locomotion tasks, which include three environments—HalfCheetah, Hopper, and Walker2d—serve as well-established benchmarks for continuous control algorithms. These domains feature high-dimensional state spaces, dense reward signals, and diverse trajectories, comprehensively evaluating policy optimization in smooth reward landscapes.

The AntMaze domain introduces a more challenging setting with sparse-reward navigation tasks that require compositional reasoning. In these tasks, an 8-degree-of-freedom ant agent must integrate suboptimal trajectory segments to achieve goal-directed navigation within maze-like environments. The evaluation includes three increasingly complex layouts—U-Maze, Medium, and Large—each presenting unique navigational challenges due to varying topological constraints and path-planning demands. This domain specifically assesses an algorithm’s ability to address credit assignment challenges and make long-horizon decisions under partial observability.

\paragraph{Baseline}
We perform a comparative evaluation of our approach in relation to several robust baseline methods, incorporating three state-of-the-art algorithms: EDP~\cite{kang2023edp}, IDQL~\cite{hansen2023idql}, Diffusion-QL~\cite{diffusion-ql}, and DTQL~\cite{chen2024DTQL}. EDP samples actions from the diffusion model by DPM-Solver~\cite{lu2022dpm-solver}. IDQL introduces a policy extraction combining the diffusion model with a reweighting scheme and  implicit Q-learning~\cite{iql}. Diffusion-QL employs a conditional diffusion model to depict the policy and leverages the Q-value function to steer the enhancement of the policy. DTQL introduces a diffusion trust region loss and aims to optimize a practical one-step policy.

\subsection{Main results on benchmark}\label{experi:main_res}
\begin{figure}[ht]
    % \centering
    \vspace{10pt}
    \hspace{-8pt}
    \includegraphics[width=\linewidth]{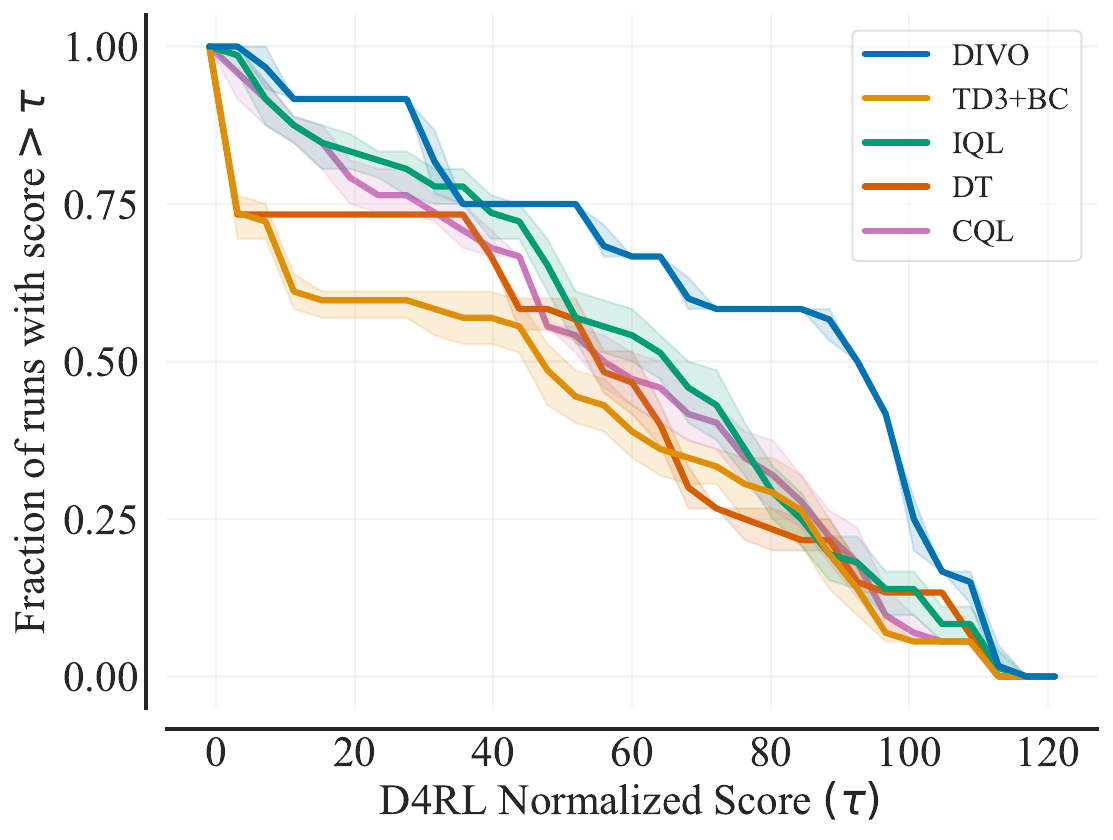}
        \caption{ To ensure reliable comparison, we evaluate performance on D4RL across 18 tasks, each with 5 random seeds.}
        \label{fig:profiles}
        % \vspace{-10pt}
    \end{figure}

This section presents the experimental evaluation of DIVO against competing baseline methods on the D4RL benchmark, with a comprehensive performance comparison summarized in Table~\ref{experi:d4rl-result}. Baseline results are directly obtained from their original publications, while our proposed DIVO is trained for 1 million environment steps using five distinct random seeds. Empirical results demonstrate that DIVO achieves state-of-the-art performance in most of the 18 benchmark tasks, significantly outperforming existing methods.
Beyond the quantitative comparisons in Table~\ref{experi:d4rl-result}, we conduct a statistically robust evaluation of several classical algorithms, with comparative analyses visualized in Figure~\ref{fig:profiles}. The convergence patterns and performance distributions in Figure~\ref{fig:profiles} provide additional insights, complementing the aggregated metrics in Table~\ref{experi:d4rl-result}. Together, these results validate the effectiveness and consistency of our approach.

 Meanwhile, the training curves of DIVO, compared to TD3+BC, CQL, IQL, and DT, are presented in Figure~\ref{fig:learning_curve}. To ensure rigorous statistical validation, we adopt the principled evaluation framework proposed by~\cite{agarwal2021deep}, incorporating robust statistical measures to account for cross-run variability. Empirical results in Figure~\ref{fig:score} consistently demonstrate the superiority of DIVO across multiple performance metrics, including higher mean and median scores, improved interquartile mean (IQM), and a reduced optimality gap compared to baseline methods. These statistically robust evaluations further substantiate the methodological advantages of our approach within the offline reinforcement learning paradigm.

\subsection{Hyperparameter Sensitivity Analysis}\label{sec:hyperpara}
This study examines the impact of hyperparameters in both the diffusion model optimization and policy improvement components, specifically $\eta$ and $\beta$. The hyperparameter $\eta$ controls the strength of the guidance for diffusion models, while $\beta$ balances policy exploration and conservatism.  
We evaluate the DIVO algorithm on the hopper-medium-v2 task, training the model for 1 million steps across five different seeds, as shown in Figure~\ref{fig:hyperany}. To analyze sensitivity to hyperparameter choices, we assess policy performance with $\eta \in \{0.5, 1.0, 1.5\}$ and $\beta \in \{0.2, 0.4, 0.6\}$. There is almost no difference in the final converged performance of these methods, which can achieve a balance between policy exploration and conservatism while ensuring effective diffusion model training. Furthermore, the consistent performance across different hyperparameter values suggests that DIVO is robust to variations in $\eta$ and $\beta$, ensuring stable policy optimization. This adaptability highlights the reliability of DIVO for diverse applications requiring consistent and effective learning dynamics.
\vspace{-5pt}

\section{Related Work}

\subsection{Generative Model for Offline RL}
Generative models have achieved significant success in offline RL. BCQ~\cite{BCQ} was the first method to employ a generative model to represent the behavior policy, constraining the learned policy to remain close to the behavior policy using conditional variational autoencoder (CVAE)\cite{sohn2015learning}. Similarly, PLAS\cite{zhou2021plas} utilizes CVAE to model the behavior policy, introducing a latent policy that naturally constrains the algorithm to stay within the dataset's support. SPOT~\cite{SPOT} pre-trains a CVAE to capture the behavior policy distribution, then uses behavior density with the CVAE to constrain the learned policy. A2PR~\cite{A2PR} utilizes a CVAE as an enhanced behavior policy and selects actions with high advantage to guide the learned policy.
Recently, with the powerful data distribution representation capabilities of diffusion models, several offline RL approaches incorporating these models have emerged. SfBC~\cite{sfbc} imitates the behavior policy using a diffusion model with score SDE~\cite{song2020score}, resampling actions from candidate actions with specific sampling weights. IDQL~\cite{hansen2023idql} adopts a similar resampling approach but imitates the behavior policy using Denoising Diffusion Probabilistic Models~(DDPM)~\cite{ho2020denoising}. Both methods require generating numerous action candidates for selection, which impedes real-world applications due to slow inference processes.
Diffusion-QL~\cite{diffusion-ql} trains a policy using a diffusion model and guides this model with a Q-value function to achieve policy improvement, comparable to TD3+BC~\cite{td3bc}. 
PSEC~\cite{liu2025psec} combines diffusion models and LoRA~\cite{hu2021lora} to achieve skill expansion and skill composition.
In contrast, DIVO employs a diffusion model to capture high-value behavior distributions from the dataset while utilizing an efficient one-step policy as the final policy. This approach effectively leverages valuable knowledge contained in the fixed dataset while avoiding the computational overhead typically associated with diffusion model inference.

\subsection{Policy Regularization for Offline RL}

% Policy regularization methods play an important role in offline RL, which usually adds a regularization term to constrain the learned policy to close the behavior policy generating the offline fixed dataset. TD3+BC~\cite{td3bc} adds a simple behavior clone term to constrain the learned policy based on the TD3~\cite{td3} framework. There are many policy regularization methods with some divergence metrics, such as Fisher divergence~\cite{Fisher-BRC}, Kullback-Leibler divergence~\cite{wu2019behavior, jaques2019way,awac}, and Maximum Mean Discrepancy (MMD)~\cite{BEAR}. AWAC uses forward
% KL to compute the policy update by sampling directly from the behavior policy. BEAR~\cite{BEAR} employs divergence regularization with MMD for policy improvement, which can be estimated based solely on samples from the distributions of the behavior policies. 
% These methods use all samples or are based on KL divergence or MMD distance to constrain the learned policy. When the fixed dataset is multimodal or heterogeneous, they are ineffective in achieving policy improvement~\cite{zhou2021plas,chen2022latent}. 
% Different from these methods, our methods learn the data distribution of the fixed dataset by the diffusion model. Meanwhile, our methods combine the value function with the diffusion model to learn an enhanced behavior policy. Then the learned policy can achieve effective policy improvement under the guidance of the diffusion model.
Policy regularization methods serve a crucial role in offline RL, typically adding a regularization term to constrain the learned policy to remain close to the behavior policy that generated the fixed offline dataset. TD3+BC~\cite{td3bc} incorporates a straightforward behavior cloning term to constrain the learned policy within the TD3~\cite{td3} framework. Various policy regularization approaches employ different divergence metrics, including Fisher divergence~\cite{Fisher-BRC}, Kullback-Leibler divergence~\cite{wu2019behavior, jaques2019way,awac}, and Maximum Mean Discrepancy (MMD)\cite{BEAR}. AWAC utilizes forward KL to compute policy updates by sampling directly from the behavior policy. BEAR\cite{BEAR} implements divergence regularization with MMD for policy improvement, which can be estimated solely from behavior policy distribution samples.
These methods either utilize all samples or rely on KL divergence or MMD distance to constrain the learned policy. However, when the fixed dataset exhibits multimodal or heterogeneous characteristics, these approaches become ineffective for achieving policy improvement~\cite{zhou2021plas,chen2022latent}.
In contrast to these methods, our approach learns the data distribution of the fixed dataset using a diffusion model. Additionally, we combine the value function with the diffusion model to learn an enhanced behavior policy. This allows the learned policy to achieve effective policy improvement under the guidance of the diffusion model.

% DOGE~\cite{} proposes to regularize the policy within the convex hull of the dataset via a state-conditioned distance function, which regresses the expectation of actions along with the given state in the offline dataset.

\section{Conclusion}
    % Conclusion
    % We introduce an innovative policy optimization approach,
    % named \textbf{DI}ffusion policies with \textbf{V}alue-conditional \textbf{O}ptimization~(DIVO) method, designed for Offline Reinforcement Learning~(RL). Unlike prior methods, DIVO models the behavior policy with a diffusion model while integrating the value function to enhance policy learning. By leveraging this value-aware diffusion model, DIVO effectively mitigates the impact of low-quality data and facilitates policy improvement. This approach introduces a more adaptive regularization mechanism, addressing the limitations of global constraints in traditional methods. Furthermore, DIVO alleviates the conservatism inherent in offline RL by selectively guiding the learned policy towards high-value actions while avoiding overfitting to suboptimal dataset samples. This enables more effective generalization and decision-making in unseen scenarios.
    % Empirical evaluations on the D4RL benchmark demonstrate that DIVO achieves SOTA performance across diverse tasks, underscoring its effectiveness in offline RL settings.
    % These results highlight DIVO as a promising direction for future advancements in offline reinforcement learning methodologies.
    We introduce an innovative policy optimization approach, named DIffusion policies with Value-conditional Optimization (DIVO) method, designed for Offline Reinforcement Learning (RL). Unlike prior methods, DIVO models the behavior policy with a diffusion model while integrating the binary-weighted mechanism to enhance policy learning. The binary-weighted mechanism in DIVO utilizes advantage values of actions in the offline dataset to guide diffusion model training, enabling more precise alignment with the dataset's distribution while selectively expanding boundaries of high-advantage actions. This approach introduces a more adaptive regularization mechanism, addressing the limitations of global constraints in traditional methods. Furthermore, DIVO alleviates the conservatism inherent in offline RL by selectively guiding the learned policy towards high-value actions while avoiding overfitting to suboptimal dataset samples, achieving a critical balance between conservatism and explorability. 
    Empirical evaluations on the D4RL benchmark demonstrate that DIVO achieves SOTA performance across diverse tasks, delivering significant improvements in average returns across locomotion tasks and outperforming existing methods in the challenging AntMaze domain with sparse rewards. These results highlight DIVO as a promising direction for future advancements in offline RL.

% \vspace{20pt}
\appendix
% \begin{appendices}
    
% \subsection{Pseudo-code}
% In this section, we provide the pseudo-code for DIVO, as presented in Table~\ref{pseudo_code}. Our algorithm is built upon the TD3+BC framework~\cite{td3bc}.

\subsection{Hyperparameter description}
In this section, we present the hyperparameter values for DIVO. The detailed hyperparameters of the algorithm are listed in Table~\ref{tab:hyperparameters}.

\begin{table}[ht]
\caption{Hyperparameter Table}
\label{tab:hyperparameters}
\small
\setlength{\tabcolsep}{5pt}
\centering
\begin{tabular}{c|l|l}
\toprule
& \textbf{Hyper-parameters} & \textbf{Value} \\
\midrule
\multirow{9}{*}{\textbf{TD3}} 
& Policy network update rate &  3e-4 \\
& {Q networks update rate} & 3e-4/1e-4 \\
% & {Q networks update rate in Antmaze} & 1e-4\\
& Count of iterations & 1e6 \\
& Target network update rate $\tau$ & 5e-3 \\
& Noise of policy & 0.2 \\
& Noise clipping of policy & (-0.5,0.5) \\
& Update frequency of policy & 2 \\
& {Discount for Mujoco} & 0.99 \\
& {Discount for Antmaze} & 0.995\\

% & &1e-4 for AntMaze \\
\hline
\multirow{11}{*}{\textbf{Network}} 
& Q networks internal layer size & 256\\
& Q network layers & 3  \\
& Q network activation function & ReLU \\
% & V-Critic hidden dim & 256  \\
% & V-Critic layers & 3  \\
% & V-Critic Activation function & ReLU \\
& Policy network internal layer size & 256  \\
& Policy network layers & 3  \\
& Policy network activation function & ReLU \\
& Mini-sample batch size & 256 \\
& Optimizer & Adam~\cite{kingma2014adam} \\
\hline
% \multicolumn{1}{c}{\bf A2PR(ours)}
\multirow{3}{*}{\textbf{DIVO}} & Normalized state & True \\
& {$\alpha$ for Mujoco} & 2.5\\ 
& {$\alpha$ for Antmaze} & \{2.5, 7.5, 20.0\}\\ 
& Diffusion Steps $K$ & 5 \\
% & &\{2.5, 7.5, 20.0\} for AntMaze \\
		\bottomrule
\hline
\end{tabular}

\end{table}

\bibliography{bibliography/bibliography}
\bibliographystyle{ieeetr}

% \newpage
% \appendix
% \subsection{First Appendix}
% \label{FirstAppendix}
 
\end{document}